 \documentclass[tablecaption=bottom]{jmlr} 

 \usepackage[export]{adjustbox}
 \usepackage{bbm}

\usepackage{booktabs}
 \usepackage{floatrow}
\usepackage{hyperref}
 \makeatletter
\def\set@curr@file#1{\def\@curr@file{#1}} 

\makeatother
\usepackage[load-configurations=version-1]{siunitx} 

\theorembodyfont{\upshape}
\theoremheaderfont{\scshape}
\theorempostheader{:}
\theoremsep{\newline}

\jmlrvolume{1}
\jmlryear{2020}
\jmlrsubmitted{March 8th}
\jmlrpublished{}

\title[MicroNet for Efficient Language Modeling]{MicroNet for Efficient Language Modeling}

 \author{\Name{Zhongxia Yan} \Email{zxyan@mit.edu}\\
 \addr Massachusetts Institute of Technology\\
  \Name{Hanrui Wang} \Email{hanrui@mit.edu}\\
  \addr Massachusetts Institute of Technology\\
  \Name{Demi Guo} \Email{dguo@college.harvard.edu}\\
  \addr Harvard University\\
  \Name{Song Han} \Email{songhan@mit.edu}\\
   \addr Massachusetts Institute of Technology}

\def\vx{\vc{x}}

\def\cL{{\cal L}}

\def\ind{\mathbbm{1}}

\newcommand{\vc}[1]{\mathbf{{#1}}}

\newcommand*{\tran}{^{\mkern-1.5mu\mathsf{T}}}

\begin{document}

\maketitle

\begin{abstract}
It is important to design compact language models for efficient deployment. We improve upon recent advances in both the language modeling domain and the model-compression domain to construct parameter and computation efficient language models. We use an efficient transformer-based architecture with adaptive embedding and softmax, differentiable non-parametric cache, Hebbian softmax, knowledge distillation, network pruning, and low-bit quantization. In this paper, we provide the winning solution to the NeurIPS 2019 MicroNet Challenge in the language modeling track. Compared to the baseline language model provided by the MicroNet Challenge, our model is 90 times more parameter-efficient and 36 times more computation-efficient while achieving the required test perplexity of 35 on the Wikitext-103 dataset. We hope that this work will aid future research into efficient language models, and we have released our full source code on \href{https://github.com/mit-han-lab/neurips-micronet}{GitHub}.
\end{abstract}

\section{Introduction}
\label{sec:intro}

Language modeling has been one of the most commonly studied sequence modeling tasks and one of the most studied tasks in the natural language processing community. Within the past few years, numerous works have improved the state-of-the-art language modeling results with deep neural network (DNN)-based sequence models. The Long Short-Term Memory (LSTM) network \citep{lstm} was designed to model dependencies in sequential data, and has successfully been applied to language modeling in a series of works, including the AWD-LSTM \citep{awdlstm}. Nevertheless, the LSTM architecture suffers from the vanishing gradient effect, limiting its ability to model long-term dependencies. On the other hand, the development of the Transformer attention architecture \citep{transformer} inspired new Transformer-based language models, such as Transformer-XL \citep{transformerxl}, which achieved new state-of-the-art in language modeling benchmarks. In addition, several works have developed architecture-agnostic enhancements such as adaptive embedding and adaptive softmax \citep{adaptive}, non-parametric cache \citep{cache}, Hebbian softmax \citep{lstmhebbiancache}, and dynamic evaluation \citep{dynamiceval}. While these works focused on improving the predictive accuracy of language models, relatively few works have optimized for parameter and computational efficiency, which are critical for tasks with hardware constraints.

On the other hand, there have been numerous advances in parameter and computation efficient neural network architectures and model-compression techniques for deep neural networks. SqueezeNet \citep{squeezenet} and MobileNet \citep{mobilenet} are efficient convolutional neural networks that take advantage of 1x1 convolutions and depthwise separable convolutions, respectively. \cite{tensorizedtransformer} applies Block-term Tensor Decomposition \citep{btd} to compress transformer-based language models. Architecture agnostic model compression techniques such as knowledge distillation \citep{distill}, network pruning \citep{deepcompression}, and trained quantization \citep{deepcompression} are commonly used techniques to either increase the predictivity of neural networks or decrease the model size. There are also recent efforts focusing on automatically designing efficient models \citep{amc, so2019evolved, apq} and designing specialized accelerators to process the compressed models \citep{scnn, sparch}. Nevertheless, these techniques are mostly developed on convolutional neural networks and computer vision tasks, and only a handful of recent works like DistilBERT \citep{distilbert} and HAT \citep{hat} applied them to natural language tasks.

In this work, we integrate advances in both the language modeling domain and the model-compression domain to construct parameter- and computation-efficient language models. Specifically, we evaluated our model with the criteria of the NeurIPS 2019 MicroNet Challenge \citep{micronet}. Compared to the baseline language model provided by the MicroNet Challenge, our model is 90 times more parameter-efficient and 36 times more computation-efficient while maintaining good performance. Our entry into the MicroNet Challenge achieved the top performance in parameter- and computation-efficiency in the language modeling track.

\section{NeurIPS 2019 MicroNet Challenge Language Modeling Track}
The NeurIPS 2019 MicroNet Challenge asks contestants to build efficient yet still performant models \citep{micronet}. In particular, the language modeling track of the competition asks participants to train efficient word-level language models on the Wikitext-103 Dataset \citep{wikitext}, a standard benchmark dataset composed of 103 million words in the training set, 217 thousand words in the validation set, and 245 thousand words in the test set, with a total of 267735 word tokens in the vocabulary. The challenge requires that entries achieve a word-level perplexity of below 35 on the test set, but otherwise does not factor perplexity into the score.

\subsection{Scoring}
Assuming an entry achieved the prerequisite perplexity threshold of 35, it is scored based on two criteria:
\begin{enumerate}
\item Parameter storage. This is the total number of 32-bit model parameters required to be loaded from disk to perform inference. Reduce-precision model parameters with less than 32-bits are counted as a fractional number. For example, an 8-bit parameter counts as $\frac{1}{4}$th of a parameter.
\item Math operations. This is the mean number of 32-bit arithmetic operations required to perform inference on each token in the tokenized version of the test set. Multiplies and additions count separately. Tokens are assumed to be fed sequentially to the model, and the model must predict the next token before receiving it. Conversion to and from reduced-precision format (e.g., int8) do not count as operations. Reduced-precision operations are counted analogously to the parameter storage.
\end{enumerate}
The total score of an entry is normalized by the size of the LSTM language model in \cite{lstmhebbiancache}, which has 159M 32-bit parameters and 318M 32-bit math operations
\[\text{Score} = \frac{\text{Parameter Storage}}{\text{159M}} + \frac{\text{Math Operations}}{\text{318M}}.\]

\section{Core Language Model}
For a corpus of tokens $\vx = (x_1,\dots,x_N)$, we take the standard approach to modeling the joint probability $P(\vx)$ by factorizing it into a product of conditional probabilities
\[P(\vx) = P(x_1) \prod_{n=2}^N P(x_n |x_{1:n-1}).\]
We model each conditional probability $P(x_n |x_{1:n-1})$ term with a transformer-based neural network with a categorical distribution output. Here discuss the set of language modeling enhancements that are applicable on top of the vanilla transformer \citep{transformer}.

\subsection{Transformer-XL Model}
The Transformer-XL model extends the vanilla Transformer decoder model \citep{transformer} by adding segment-level recurrence with state reuse and relative positional embeddings \citep{transformerxl}. These modifications improve the inference-time computational efficiency greatly when comparing with the vanilla Transformer. To maintain a fixed context size for predicting each token, the vanilla transformer requires full forward-computation of the entire context due to its absolute positional embedding, while the Transformer-XL merely needs to compute the forward-computation for one token due to its relative positional embeddings. We note this computational advantage of the Transformer-XL and incorporate it into our model.

\subsection{Joint Optimization of Groups of Short Contexts}
\label{subsec:context}
While Transformer-XL focuses on modeling long-term dependencies with per-layer context size $C_{xl}$ on the order of thousands of tokens, we experiment with using \textit{short} per-layer context size $C \ll C_{xl}$ in the MicroNet challenge; this approach allows us to train \textit{multiple} contexts jointly and reduces the test-time computation cost.

Transformer-XL and our model both consist of $L$ layers, allowing the size of the context to grow linearly with the number of layers. However, due to limitations of GPU memory, Transformer-XL only backpropagates to the previous $C_{xl} - 1$ dependencies at training-time---dropping the gradients to further dependencies---despite using $L(C_{xl} - 1) + 1$ previous dependencies at test-time. In addition, \cite{transformerxl} reported diminishing decreases in perplexity as $C_{xl}$ increases.

In contrast, we choose a relatively short per-layer context size $C$ on the order of a few hundred tokens but jointly optimize over an \textit{extended context} of $C_e$ tokens. Each additional layer allows us to backpropagate to $C - 1$ \textit{more} dependencies at training-time, so in total we backpropagate to dependencies up to $\min(C_e, L(C - 1) + 1)$ tokens away across $L$ layers. In practice we choose $C_e \geq L(C - 1) + 1$ so that we can backpropagate to up to $L(C-1) + 1$ dependencies at training-time. At test-time, each prediction has an efficient per-layer context of $C$ and a total context of $L(C-1) + 1$. In Appendix~\ref{app:context}, we verify that our choice of $C \ll C_{xl}$ is ideal in parameter- and computation-efficient settings. We illustrate joint training across short contexts in \figureref{fig:transformers}.
\begin{figure}[htbp]
  \floatconts
    {fig:transformers}
    {\caption{Transformer-XL vs. our model. Black lines represent forward- and backward-propagation while gray lines represent forward-propagation only. Each blue triangle represents all the dependencies of the top node in the triangle.}}
    {%
      \subfigure[Two consecutive training steps for Transformer-XL: each step the model predicts $C_{xl} - 1$ new tokens from the computed dependencies (red) and the stored dependencies (yellow). Gradient is only backpropagated to the computed dependencies but not the stored dependencies.]{\label{fig:transformer_xl}
        \def\svgwidth{0.45\linewidth}
\begingroup%
  \makeatletter%
  \providecommand\color[2][]{%
    \errmessage{(Inkscape) Color is used for the text in Inkscape, but the package 'color.sty' is not loaded}%
    \renewcommand\color[2][]{}%
  }%
  \providecommand\transparent[1]{%
    \errmessage{(Inkscape) Transparency is used (non-zero) for the text in Inkscape, but the package 'transparent.sty' is not loaded}%
    \renewcommand\transparent[1]{}%
  }%
  \providecommand\rotatebox[2]{#2}%
  \newcommand*\fsize{\dimexpr\f@size pt\relax}%
  \newcommand*\lineheight[1]{\fontsize{\fsize}{#1\fsize}\selectfont}%
  \ifx\svgwidth\undefined%
    \setlength{\unitlength}{85.5bp}%
    \ifx\svgscale\undefined%
      \relax%
    \else%
      \setlength{\unitlength}{\unitlength * \real{\svgscale}}%
    \fi%
  \else%
    \setlength{\unitlength}{\svgwidth}%
  \fi%
  \global\let\svgwidth\undefined%
  \global\let\svgscale\undefined%
  \makeatother%
  \begin{picture}(1,0.55263158)%
    \lineheight{1}%
    \setlength\tabcolsep{0pt}%
    \put(0,0){\includegraphics[width=\unitlength,page=1]{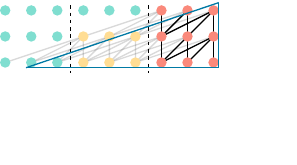}}%
    \put(0.44,0.31){\makebox(0,0)[lt]{\lineheight{1.25}\smash{\begin{tabular}[t]{l}$\underbrace{\hspace{2.1cm}}_{C_{xl} = 4}$\end{tabular}}}}%
    \put(0.74,0.425){\makebox(0,0)[lt]{\lineheight{1.25}\smash{\begin{tabular}[t]{l}$\left.\rule{0cm}{0.9cm}\right\} L = 2$\end{tabular}}}}%
    \put(0,0){\includegraphics[width=\unitlength,page=2]{figures/transformer_xl.pdf}}%
  \end{picture}%
\endgroup%
}
        \hspace{0.05\textwidth}
        \subfigure[One training step for our model: the model predicts $C_e$ new tokens by computing all dependencies and keeping all gradient information. In the blue triangle, we backpropagate the gradient of the top node to all of its $L(C - 1) + 1$ dependencies.]{\label{fig:transformer_ours}%
        \def\svgwidth{0.45\linewidth}
\begingroup%
  \makeatletter%
  \providecommand\color[2][]{%
    \errmessage{(Inkscape) Color is used for the text in Inkscape, but the package 'color.sty' is not loaded}%
    \renewcommand\color[2][]{}%
  }%
  \providecommand\transparent[1]{%
    \errmessage{(Inkscape) Transparency is used (non-zero) for the text in Inkscape, but the package 'transparent.sty' is not loaded}%
    \renewcommand\transparent[1]{}%
  }%
  \providecommand\rotatebox[2]{#2}%
  \newcommand*\fsize{\dimexpr\f@size pt\relax}%
  \newcommand*\lineheight[1]{\fontsize{\fsize}{#1\fsize}\selectfont}%
  \ifx\svgwidth\undefined%
    \setlength{\unitlength}{63bp}%
    \ifx\svgscale\undefined%
      \relax%
    \else%
      \setlength{\unitlength}{\unitlength * \real{\svgscale}}%
    \fi%
  \else%
    \setlength{\unitlength}{\svgwidth}%
  \fi%
  \global\let\svgwidth\undefined%
  \global\let\svgscale\undefined%
  \makeatother%
  \begin{picture}(1,0.33333333)%
    \lineheight{1}%
    \setlength\tabcolsep{0pt}%
    \put(0,0){\includegraphics[width=\unitlength,page=1]{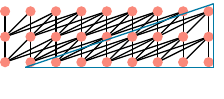}}%
    \put(0.56,0.135){\makebox(0,0)[lt]{\lineheight{1.25}\smash{\begin{tabular}[t]{l}$\underbrace{\hspace{2.8cm}}_{C = 4}$\end{tabular}}}}%
    \put(0.0,0.06){\makebox(0,0)[lt]{\lineheight{1.25}\smash{\begin{tabular}[t]{l}$\underbrace{\hspace{6.8cm}}_{C_e = 9}$\end{tabular}}}}%
  \end{picture}%
\endgroup%
}
    }
  \end{figure}

\subsection{Adaptive Embedding and Softmax}
We use the adaptive embedding and adaptive softmax as presented in \cite{adaptive}. Naive word embedding and softmax require parameter, memory, and computation costs proportional to the vocabulary size, which significantly reduces the batch size available for tasks like Wikitext-103, which has a vocabulary size of 267735. Intuitively, adaptive embedding and softmax allow us to allocate more representational power to the embedding space of tokens that occur more frequently in the training set; we sort the tokens in the vocabulary by frequency and assign smaller embedding vectors to less frequent vocabulary.

\subsection{Hebbian Updates}
We experiment with using Hebbian softmax \citep{lstmhebbiancache}, which updates the output embedding by combining traditional gradient descent and interpolation with the last hidden activations. \cite{lstmhebbiancache} suggests that non-parametric Hebbian updates help the embedding memorize the representations of infrequent tokens.

\subsection{Differentiable Non-parametric Cache}
\label{subsec:cache}
We explore the use of a non-parametric cache with a cache size on the order of thousands of activations. As described by \cite{cache}, the non-parametric cache stores the most recent activations of the network and their corresponding labels, which improves prediction for infrequent tokens that appeared in the recent context. We define cache size $n_\text{cache}$ and model activation $h$. To calculate cross entropy loss for token $x_n$, we use predicted token probability $P(x_n)$ given by
\begin{align*}
  P(x_n|x_{1:n-1}) &= (1 - \lambda_\text{cache}) P_\text{softmax}(x_n|x_{1:n-1}) + \lambda_\text{cache} P_\text{cache}(x_n|x_{1:n-1})\\
  P_\text{cache}(x_n|x_{1:n-1}) &\propto \sum_{i = 1}^{n_\text{cache}} e^{\theta_\text{cache} h_{n - i}\tran h_n}\ind_{x_{n - i} = x_n},
\end{align*}
where $P_\text{softmax}$ is the output probability of the language model and $P_\text{cache}$ is the binned softmax of cache similarities.

While \cite{cache} and \cite{lstmhebbiancache} perform hyperparameter grid-search over $\theta_\text{cache}$ and $\lambda_\text{cache}$ on the validation set, we make an enhancement to jointly optimize the cache hyperparameters with gradient descent during training. We make an additional observation that batches at training-time are randomly sampled with no relation to the previous batch, while batches at test-time are iterated in order. To exploit this structure, we explore the effect of local search on the cache parameters $\theta_\text{cache}$ and $\lambda_\text{cache}$ on the validation set \textit{after} training. Since we may use a different cache size for local search, we define $n_\text{cache}^\text{t}$ and $n_\text{cache}^\text{s}$ to denote the cache size at training-time and local search-time, respectively.

\section{Compression Techniques}
We experiment with three compression techniques---knowledge distillation, pruning, and quantization---on top of our core language model.

\subsection{Knowledge Distillation}
The knowledge distillation technique \citep{distill} first learns a larger, more predictive teacher model $P_\text{teacher}$ then transfers its behavior into a more compact student model. For efficiency, we compute the largest 30 ``soft'' labels $S_{30} \subset \{P_\text{teacher}(x|x_{1: n-1}) : x\in \text{Vocab}\}$ for token $x_n$ in the training set. We train the student model with distillation loss $\cL_\text{soft}$ in addition to the standard crossentropy loss $\cL_\text{hard}$, which uses $x_n$ as the ground truth label. We apply teacher annealing \citep{distillanneal} to linearly reduce the weight of the $\cL_\text{soft}$ from $\lambda_\text{max}$ to $\lambda_\text{min}$ in $T$ training steps. The total loss for token $x_n$ at step $t$ is
\begin{align*}
  \cL_t(x_n) &= (1 - \lambda_\text{soft}(t)) \cL_{\text{hard}}(x_n) + \lambda_\text{soft}(t) \cL_{\text{soft}}(P_\text{teacher})\\
  \cL_{\text{soft}}(P_\text{teacher}) &= -\sum_{x \in S_{30}} P_{\text{teacher}}(x|x_{1: n-1}) \log P(x|x_{1: n-1})\\
  \lambda_\text{soft}(t) &= \lambda_\text{max} - \frac{t}{T} (\lambda_\text{max} - \lambda_\text{min})
\end{align*}

\subsection{Pruning}
To prune a given model, we perform sensitivity analysis \citep{sensitivity} on the model's parameter matrices $\Phi = \{\phi_1, \dots, \phi_p\}$ to analyze how the performance of the model degrades with sparsity $\rho_\phi$ of each parameter. We empirically obtain the $\xi_\phi = f_\phi(\rho_\phi)$, which maps the sparsity of $\phi$ to the model perplexity $\xi_\phi$. Next, we choose a target model sparsity $\rho$ as a hyperparameter, which is a weighted combination of the parameter sparsities $\rho_\phi$.
\[\rho = \frac{\sum_{\phi \in \Phi} |\phi|\rho_\phi}{\sum_{\phi \in \Phi} |\phi|} = \frac{\sum_{\phi \in \Phi} |\phi|f_\phi^{-1}(\xi^*)}{\sum_{\phi \in \Phi} |\phi|} \]
In practice, we solve for the perplexity threshold $\xi^*$ by doing a binary search on values of $\xi^*$ until we achieve a model sparsity of $\rho$. We then use Automatic Gradual Pruning \citep{agp} to simultaneously prune parameters with target sparsities $\rho_\phi = f_\phi(\xi^*)$, which specifies different levels of aggressiveness for different parameters. We explore the effect on model size and performance for several values of $\rho$.

\subsection{Quantization}
To quantize a given model, we choose a bit-width $w$ to quantize our model to, then perform one step of quantization-aware training with linear-range symmetric fake-quantization \citep{qat} on the model parameters and activations. To ensure that our model can be expressed in reduced-precision integer representation, we make sure that the last $w$ bits of the mantissa of the inverse scale factors are $0$. We perform quantization only on the non-embedding weights and on activations with no normalization operations; we do not quantize the output of layer normalization and softmax layers. Due to competition rules, we assume that all addition operations take place as 32-bit operations.

\section{Experiments}
\begin{figure}[htbp]
  \includegraphics[width=0.5\linewidth]{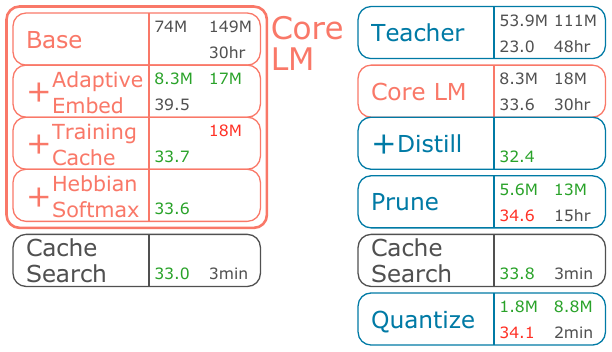}
  \caption{Our pipeline without (left) and with (right) compression techniques, using the hyperparameters in Appendix~\ref{app:hyperparameters}. From top to bottom, each stack displays the progression of techniques. Each row displays associated metrics: parameters (top left), operations (top right), validation perplexity (bottom left), and estimated processing time (bottom right). Metrics are displayed when changed from the previous row, with green for desirable change and red for undesirable. Red rows represent Core LM techniques, blue rows represent compression techniques, and gray rows represent cache search; joined rows represent joint training.}
  \label{fig:pipeline}
\end{figure}

We discuss experiments on the effect of the training cache size $n_\text{cache}^\text{t}$ and the effects of compression techniques. Due to space constraint, we defer the detailed list of model configurations, comparison with state-of-the-art language models, analysis of per-layer context size, ablation study on the training cache sizes, ablation study on Hebbian softmax, and analysis of search cache sizes to Appendices \ref{app:hyperparameters}, \ref{app:sota}, \ref{app:context}, \ref{app:cache_train}, \ref{app:hebbian}, and \ref{app:cache_search}, respectively. We depict our overall pipeline in \figureref{fig:pipeline} with performances results and processing time estimates.

\subsection{Overall Results}
We report our performance with the hyperparameters detailed in Appendix~\ref{app:hyperparameters}. Without compression techniques, our core language model (Core LM) achieves a validation perplexity of 33.6 with 8.3M parameters and 18M math operations. Adding the compression techniques, our model achieves a validation perplexity of 34.1 and test perplexity of 35.0, using 1.8M 32-bit parameters and 8.8M 32-bit math operations for a final MicroNet score of 0.0387. This is equivalent to a 90-fold reduction in parameter size and a 36-fold reduction in math operations compared to the MicroNet baseline. We compare our results to those of other MicroNet challenge language modeling track participants in \tableref{tab:micronet}.

\begin{table}[htbp]
  \floatconts
    {tab:micronet}
    {\caption{Performance of teams in the MicroNet Challenge Language Modeling track \citep{micronet}. Ours$^\dagger$ is our original submission to the challenge, which contains an evaluation error. Ours$^*$ is the work presented in this paper with the error resolved. Clova AI / Kyoto University does not report methods. JAIST / ISM uses a QRNN base-model \citep{qrnn} rather than a Transformer base-model.}}%
    {%
      \begin{tabular}{|l|l|l|l|}
      \hline
      \bfseries Participant & \bfseries MicroNet Score \\
      \hline
      Ours$^*$ & 0.0387\\
      MIT-HAN-Lab (Ours$^\dagger$) & 0.0475\\
      Clova AI / Kyoto University & 0.1657\\
      JAIST / ISM & 0.8232\\\hline
      \end{tabular}
    }
  \end{table}

\subsection{Cache Effect Analysis}
\begin{figure}[htbp]
  \floatconts
    {fig:cache_losses}
    {\caption{Blue: cumulative sum over the binned validation loss-difference between $n_\text{cache}^\text{t} \in \{\text{no-cache}, 1000\}$ and $n_\text{cache}^\text{t} = 2000$. Red: cumulative bin size for comparison. If the loss-difference were equally distributed across all tokens, the cumulative loss-difference would exactly match the cumulative bin size. All experiments are performed with $C = 97$, Hebbian softmax, and no compression techniques.}}
    {%
      \subfigure[Binned by token index]{
        \label{fig:token_loss}%
        \includegraphics[width=0.5\linewidth]{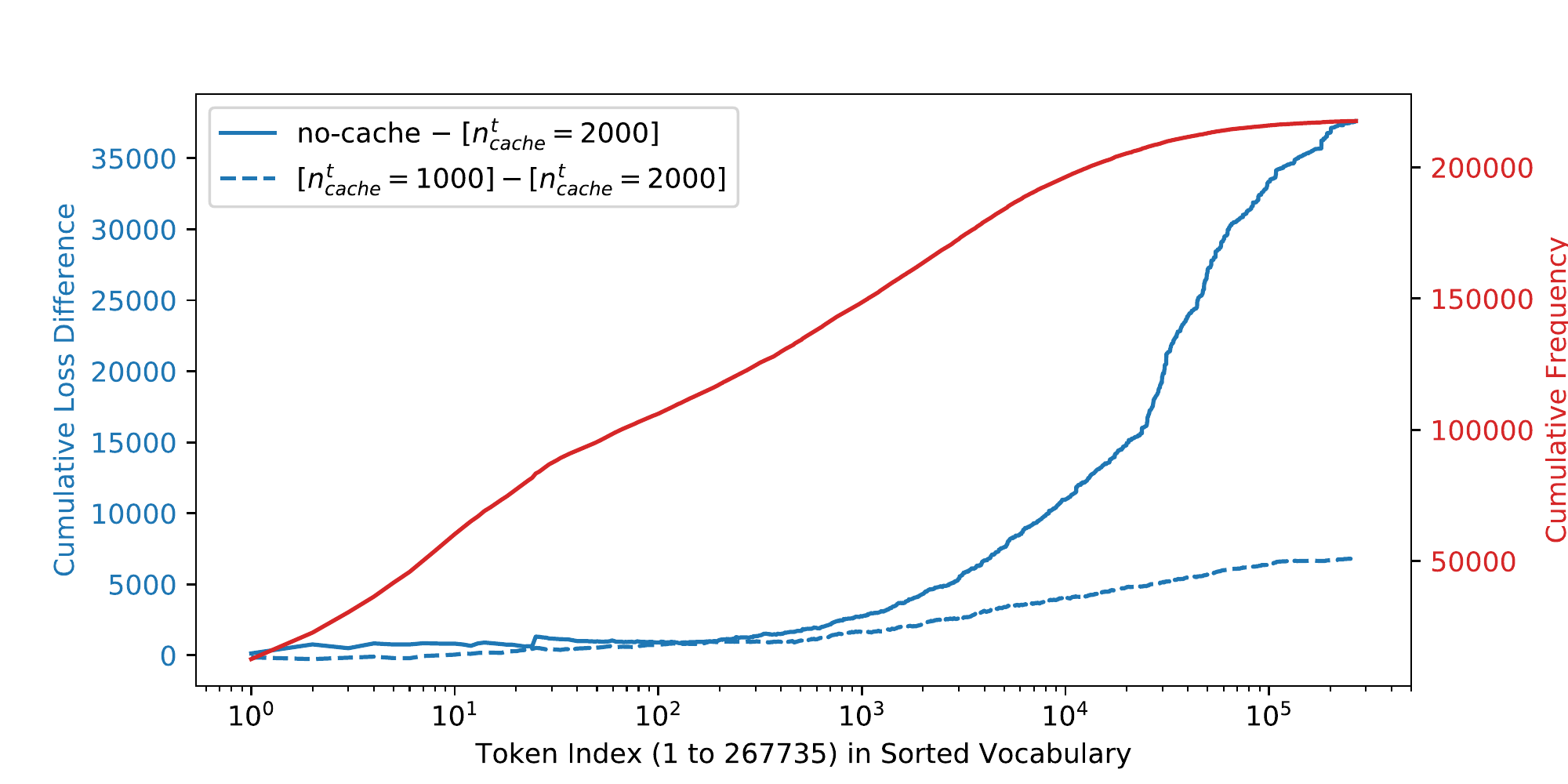}}%
      \subfigure[Binned by token gap]{
        \label{fig:gap_loss}%
        \includegraphics[width=0.5\linewidth]{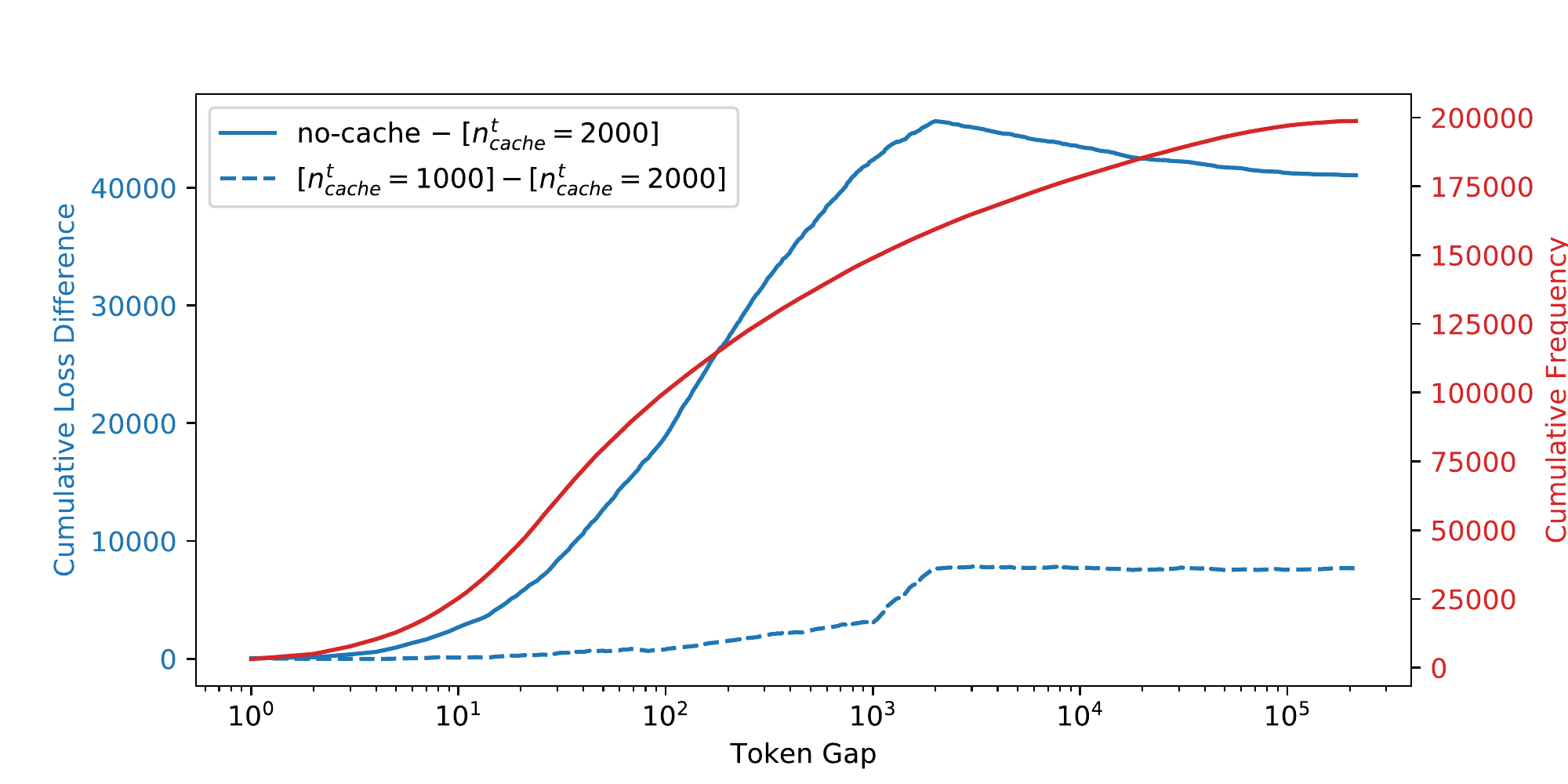}}
    }
  \end{figure}

We analyze the token-level effect of the non-parametric cache on different token bins within the validation loss. \cite{cache} states that the cache increase prediction accuracy on rare labels within the cache range. We analyze three models with training cache sizes $n_\text{cache}^\text{t} \in \{\text{no-cache}, 1000, 2000\}$. In \figureref{fig:token_loss}, we bin the loss by token index sorted in decreasing order of token frequency and compare loss incurred in $n_\text{cache}^\text{t} \in \{\text{no-cache}, 1000\}$ to loss incurred in $n_\text{cache}^\text{t} = 2000$. We observe that most of the additional loss in $n_\text{cache}^\text{t} \in \{\text{no-cache}, 1000\}$ is attributed to rare tokens, with token indices $> 10^3$ in the vocabulary.

In \figureref{fig:gap_loss}, we define the token gap to be the distance between a token and the previous occurrence of that token in the validation set. Similarly to above, we bin the loss by token gap and identify which token gap incur additional loss. We observe that $n_\text{cache}^\text{t} = 2000$ significantly outperforms no-cache on tokens with gaps between $100$ and $2000$ and slightly underperforms no-cache on tokens with gaps $\geq 2000$. We also observe that $n_\text{cache}^\text{t} = 2000$ outperforms $n_\text{cache}^\text{t} = 1000$ on tokens with gaps between $1000$ and $2000$. In both cases, the non-parametric cache improves prediction accuracy within the cache range.

\subsection{Effect of Distillation, Pruning, and Quantization}
We take per-layer context $C \in \{97, 129, 257\}$ and run an ablation study on the compression techniques, comparing the performance at each stage of compression to the performance of the Core LM. We perform cache search \textit{before} quantization because we empirically found that quantization immobilizes the cache parameters, and cache search after quantization fails to yield better parameters. Consistent with Appendix~\ref{app:context}, we observe that $C \in \{129, 257\}$ results in significantly lower perplexity than $C = 97$ for much of the pipeline before quantization. Surprisingly, the perplexities of $C \in \{129, 257\}$ increases significantly after quantization, while the perplexity of $C = 97$ only increases marginally. We suspect that this phenomenon is due to the fact that $C = 97$ is more reliant on the cache; $C = 97$ has cache weight $\lambda_\text{cache} = 0.15$, larger than $0.13$ for $C \in \{129, 257\}$. Overall, we find $[C = 97, \rho = 0.358, n_\text{cache}^\text{s} = 3000, \text{bits} = 9]$ to be the best hyperparameters for the MicroNet challenge.

\begin{table}[htbp]
  \floatconts
    {tab:compression}
    {\caption{Performance of different choices of per-layer context $C$, model sparsity $\rho$, local search cache size $n_\text{cache}^\text{s}$, and quantization bit. We bold the overall best result for the MicroNet challenge. ``Val'' and ``Test'' denote respective perplexities.}}%
    {%
      \begin{tabular}{|cc|c|cc|cc|ccccc|}
      \hline
      \multicolumn{2}{|c|}{\bfseries Core LM} & \bfseries +Distill & \multicolumn{2}{|c|}{\bfseries +Prune} & \multicolumn{2}{|c|}{\bfseries +Search} & \multicolumn{5}{|c|}{\bfseries +Quantize}\\
      \hline
      \bfseries $C$ & Val & Val & $\rho$ & Val & $n_\text{cache}^\text{s}$ & Val & bits & Params & Ops & Val & Test\\
      \hline
      \bfseries 97 & \bfseries 33.6 & \bfseries 32.9 & 0.239 & 33.6 & 2000 & 33.2 & 8 & 1.8M & 9.5M & 34.7 & 35.6\\
      \cline{8-12}
      & & & & & & & 9 & 2.0M & 9.7M & 33.5 & 34.4\\
      \cline{4-12}
      & & & \bfseries 0.358 & \bfseries 34.6 & 2000 & 34.1 & 9 & 1.8M & 8.5M & 34.5 & 35.3\\
      \cline{8-12}
      & & & & & & & 10 & 1.9M & 8.7M & 34.2 & 35.1\\
      \cline{6-12}
      & & & & & \bfseries 3000 & \bfseries 33.8 & 8 & 1.6M & 8.6M & 35.2 & 36.1\\
      \cline{8-12}
      & & & & & & & \bfseries 9 & \bfseries 1.8M & \bfseries 8.9M & \bfseries 34.1 & \bfseries 35.0\\
      \cline{6-12}
      & & & & & 4000 & 33.6 & 8 & 1.6M & 8.9M & 35.1 & 36.0\\
      \cline{8-12}
      & & & & & & & 9 & 1.8M & 9.1M & 34.0 & 34.8\\
      \cline{4-12}
      & & & 0.477 & 36.2 & 2000 & 35.7 & 9 & 1.6M & 7.5M & 36.1 & 36.8\\
      \cline{6-12}
      & & & & & 4000 & 35.2 & 10 & 1.7M & 8.4M & 35.3 & 36.0\\
      \hline
      129 & 33.3 & 32.7 & 0.358 & 34.0 & 2000 & 33.5 & 9 & 1.8M & 8.6M & 35.0 & 36.0\\
      \cline{8-12}
      & & & & & & & 10 & 1.9M & 8.8M & 33.8 & 34.8\\
      \cline{6-12}
      & & & & & 3000 & 33.2 & 9 & 1.8M & 8.9M & 34.7 & 35.7\\
      \hline
      257 & 32.2 & 31.6 & 0.358 & 33.3 & 2000 & 32.9 & 9 & 1.8M & 9.1M & 34.3 & 35.2\\
      \hline
      \end{tabular}
    }
  \end{table}

\section{Conclusions}
In our work, we combined transformer-based methods and cache-based language modeling methods to significantly reduce the total amount of parameters and computation while maintaining good perplexity. Furthermore, we showed that model-compression techniques achieve large reductions in parameter size and computation. We hope that this work will aid future research into efficient language models, and we have released our full source code on \href{https://github.com/mit-han-lab/neurips-micronet}{GitHub}.

\section*{Acknowledgments}
We thank Facebook Faculty Award, AWS Machine Learning Award, and AMD for sponsoring this research. We are grateful to Phillip Isola for helpful discussions.

\newpage
\bibliography{micronet.bib}

\newpage
\appendix
\section*{Appendices}
\addcontentsline{toc}{section}{Appendices}
\renewcommand{\thesubsection}{\Alph{subsection}}

\subsection{Configurations and Hyperparameters}
\label{app:hyperparameters}
Our best model use the following hyperparameters. Unless otherwise stated, experiments discussed in this paper also use these hyperparameters. The full implementation can be found in the code release.
\begin{enumerate}
  \item Adaptive embedding and softmax: after sorting in order of increasing frequency, we divide the vocabulary into three bins: $[1, 3500], [3501, 25000], [25001, 267735]$. For the tokens in each bin, we use embedding vectors of sizes $[256, 64, 4]$, respectively.
  \item Base transformer: we use a base transformer with context size $C = 97$, extended context $C_e = 1152$, $L = 8$ layers, input and output dimension $d_\text{model} = 256$, $h = 8$ attention heads, key and value dimensions $d_k = d_v = 24$, and an ``inner'' fully connected layer dimension $d_{\textit{ff}} = 768$.
  \item Cache training: we use a cache of size $n_\text{cache}^\text{t} = 2000$ with initial parameters $\theta_{\text{cache},0} = 0.016$ and $\lambda_{\text{cache},0} = 0.07$.
  \item Hebbian softmax: we use a minimum discount factor $\gamma_\text{hebbian} = 0.01$ and a smoothing limit $T_\text{hebbian} = 500$ classes.
  \item Training: we train for $T = 200000$ steps with the Adam Optimizer \citep{adam}, learning rate $0.0001$, and cosine learning rate decay with $1000$ linear-warmup steps. We choose the largest batch size that fits in memory.
  \item Distillation: we distill from a teacher model trained with $L = 16$, embedding vectors of size $[512, 256, 16]$, $d_\text{model} = 512$, $d_k = d_v = 64$, $d_{\textit{ff}} = 1536$, learning rate $0.0005$, dropout $0.1$, no cache, and otherwise the same hyperparameters as the student model. The teacher model has 53.8M parameters and a validation perplexity of 23.0. We train the student model with $\lambda_\text{max} = 0.5$ and $\lambda_\text{min} = 0.05$.
  \item Pruning: we choose a target sparsity $\rho = 0.358$. We initialize the model with the best model from the distillation step, prune this model to an initial sparsity $\rho_0 = 0.16$, then train for $175000$ steps, pruning every $1000$ steps. We use the Distiller \citep{distiller} implementation of the Adaptive Gradual Pruning algorithm.
  \item Cache local search: we perform local search on the cache parameters $\theta_\text{cache}$ and $\lambda_\text{cache}$ with a search cache size $n_\text{cache}^\text{s} = 3000$.
  \item Quantization: we quantize the model to 9 bits using the Distiller \citep{distiller} implementation of quantization-aware training and symmetric linear-range quantization.
\end{enumerate}
We train the large teacher model on eight GeForce RTX 2080 Ti GPUs and all other models on a single GeForce GTX Titan X GPU, RTX 2080 Ti, or V100 GPU.

\subsection{Comparisons to SOTA}
\label{app:sota}
In \tableref{tab:compare}, we compare the performance of our models against performances of state-of-the-art methods reported by \cite{deq}. These methods use full word embeddings with embedding dimension $512$, while we use embedding dimension $256$ to reduce parameters. We compare a base-model with full word embedding and also compare our best small adaptive embedding models (with and without cache). We find that our base-model performance is comparable to Transformer-XL despite requiring half as many embedding parameters. Adding adaptive embedding and cache allows us to stay competitive with state-of-the-art methods while using much fewer parameters.

\begin{table}[htbp]
\floatconts
  {tab:compare}
  {\caption{Our models are competitive with SOTA transformer-based models while using much less embedding parameters. Here we do not use Hebbian softmax. We note that the DEQ Transformer \cite{deq} uses root-finding methods to optimize model parameters and does not provide an estimate on computations required.}}%
  {%
    \begin{tabular}{|l|l|l|l|l|l|}
    \hline
    \abovestrut{2.2ex}\bfseries Model & \bfseries Params & \bfseries Embedding & \bfseries Model & \bfseries $n_\text{cache}^\text{t}$ & \bfseries Test PPL\\
     &  & \bfseries Params & \bfseries Params &  & \\
    \hline
    Transformer-XL Small & 139M & 134M & 4.9M & 0 & 35.8\\
    DEQ-Transformer Small & 138M & 134M & 4.5M & 0 & 32.4\\
    Ours, Base & 73.6M & 68.5M & 5.0M & 0 & 36.5\\
    Ours, Adaptive & 8.3M & 3.3M & 5.0M & 0 & 41.3\\
    Ours, Adaptive, Cache & 8.3M & 3.3M & 5.0M & 2000 & 34.9\\
    \hline
    \end{tabular}
  }
\end{table}

\subsection{Context Length Experiment}
\label{app:context}
We perform a sweep over per-layer context length $C$ as detailed in \tableref{tab:context} and analyze model performance and math operations. We see that using $C = 257$ or $C = 129$ achieves a good trade-off between the number of operations and the validation perplexity. Interestingly, $C \in \{513, 1025\}$ results in both increasing computation cost and worse perplexity; we suspect that the latter may be due to the small model size that we use.
\begin{table}[htbp]
  \floatconts
    {tab:context}
    {\caption{Performance of different per-layer context $C$ with only adaptive softmax; we do not use cache, Hebbian softmax, or compression techniques.}}%
    {%
      \begin{tabular}{|l|l|l|l|}
      \hline
      \bfseries Context $C$ & \bfseries Params & \bfseries Operations & \bfseries Val PPL\\
      \hline
      1025 & 8.5M & 23.4M & 38.4\\
      513 & 8.4M & 20.1M & 37.7\\
      257 & 8.3M & 18.5M & 37.5\\
      129 & 8.3M & 17.6M & 38.6\\
      97 & 8.3M & 17.4M & 39.5\\
      65 & 8.3M & 17.2M & 40.5\\
      \hline
      \end{tabular}
    }
  \end{table}

\subsection{Training Cache Size Ablation Study}
\label{app:cache_train}
We conduct an ablation study on the effect of the training cache size on perplexity. In \tableref{tab:cache}, we see that the non-parametric cache is crucial to good performance; the 8.3M parameter model equiped with $n_\text{cache}^{t} = 2000$ performs similarly to a model with between 11M and 15M parameters with no cache.

\begin{table}[htbp]
  \floatconts
    {tab:cache}
    {\caption{Non-parametric cache improves the performance of the 8.3M parameter model greatly at little extra cost. We use $C = 97$ and Hebbian softmax for all models, but do not apply local cache search or compression techniques.}}%
    {%
      \begin{tabular}{|l|l|l|l|}
      \hline
      \abovestrut{2.2ex}\bfseries $n_\text{cache}^\text{t}$ & \bfseries Params & \bfseries Operations & \bfseries Val PPL\\\hline
      0 & 8.3M & 17.4M & 39.2\\
      0 & 11.0M & 23.3M & 37.6\\
      0 & 15.2M & 31.6M & 32.4\\
      1000 & 8.3M & 17.9M & 34.8\\
      2000 & 8.3M & 18.4M & 33.6\\\hline
      \end{tabular}
    }
  \end{table}

\subsection{Hebbian Softmax Ablation Study}
\label{app:hebbian}
We perform an ablation study on the Hebbian softmax technique presented in \cite{lstmhebbiancache}. Unlike in their work, we do not see a significant decrease in perplexity when using the Hebbian softmax technique. A potential explanation is that they use an LSTM base-model and regular softmax, whereas we use a transformer base-model and adaptive softmax.

\begin{table}[htbp]
  \floatconts
    {tab:hebbian}
    {\caption{Comparison of validation perplexity with and without Hebbian softmax in our model vs in \cite{lstmhebbiancache}. We use context $C = 97$ for all experiments here. For our experiments with trained cache, we use $n_\text{cache}^\text{t} = 2000$.}}%
    {%
      \begin{tabular}{|l|l|l|l|l|}
      \hline
      \abovestrut{2.2ex} &  & \bfseries No Cache & \bfseries With Trained Cache\\\hline
      \cite{lstmhebbiancache} & No Hebbian & 36.0 & 34.5 \\
       & With Hebbian & 34.1 & 29.7\\\hline
      Ours & No Hebbian & 39.5 & 33.7\\
      \belowstrut{0.2ex} & With Hebbian & 39.2 & 33.6\\\hline
      \end{tabular}
    }
  \end{table}

\subsection{Cache Training and Cache Search Experiment}
\label{app:cache_search}
As mentioned in Subsection~\ref{subsec:cache}, we sample random extended contexts $C_e$ at training-time but sequentially iterate over contexts in the validation and test sets; the latter setup allows for stronger temporal locality. We therefore expect that even if we train cache weight $\lambda_\text{cache}$ at training-time, increasing this hyperparameter yields better performance at test-times. We empirically observe that the validation perplexity is a smooth function of $\theta_\text{cache}$ and $\lambda_\text{cache}$ with a clear global minimum, so we perform local search on $\theta_\text{cache}$ and $\lambda_\text{cache}$ to minimize the validation perplexity, instead of using grid-search. In \tableref{tab:cache_search}, we compare the performance of combinations of training cache size and search cache size, which we denote as $(n_\text{cache}^\text{t}, n_\text{cache}^\text{s})$. Interestingly, $(2000, \text{no-search})$ significantly outperforms $(\text{no-cache}, 2000)$, which suggests that jointly training of the cache hyperparameters is preferable to solely performing hyperparameter search over the validation set. We suspect that joint training learns more cache-friendly final layer activations $h$. We also observe that performing local search \textit{after training} still improves performance, with bigger improvements for larger $n_\text{cache}^\text{s}$. Correspondingly, we observe that the cache weight $\lambda_\text{cache}$ increases with $n_\text{cache}^\text{s}$.

\begin{table}[htbp]
  \floatconts
    {tab:cache_search}
    {\caption{Performance of models with different combinations of training and local search cache sizes. Note that $(2000, 2000)$, $(2000, 4000)$, $(2000, 6000)$, and $(2000, 8000)$ merely performs local search on the trained model from $(2000, \text{no-search})$. We use $C = 97$ and Hebbian softmax for all models.}}%
    {%
      \begin{tabular}{|l|l|l|l|l|}
      \hline
      \bfseries Training $n_\text{cache}^\text{t}$ & \bfseries Search $n_\text{cache}^\text{s}$ & \bfseries Val PPL & \bfseries $\lambda_\text{cache}$ & \bfseries $\theta_\text{cache}$ \\\hline
      $\text{no-cache}$ & $2000$ & 34.8 & $0.0922$ & $0.0197$\\
      $\text{no-cache}$ & $4000$ & 34.7 & $0.0964$ & $0.0213$\\
      $\text{no-cache}$ & $6000$ & 34.8 & $0.0963$ & $0.0221$\\
      $2000$ & $\text{no-search}$ & 33.6 & $0.0689$ & $0.0242$\\
      $2000$ & $2000$ & 33.0 & $0.145$ & $0.0266$\\
      $2000$ & $4000$ & 32.6 & $0.156$ & $0.0276$\\
      $2000$ & $6000$ & 32.4 & $0.158$ & $0.0281$\\
      $2000$ & $8000$ & 32.4 & $0.160$ & $0.0284$\\
      \hline
      \end{tabular}
    }
  \end{table}
\end{document}